# Probability Reversal and the Disjunction Effect in Reasoning Systems

Subhash Kak

**Abstract.**
Data based judgments that go into artificial intelligence applications may undergo paradoxical reversal when seemingly unnecessary additional data is provided. Examples of this are Simpson's reversal and the disjunction effect where the beliefs about the data change once it is presented or aggregated differently. Sometimes the significance of the difference can be evaluated using statistical tests such as Pearson's chi-squared or Fisher's exact test, but this may not be helpful in threshold-based decision systems that operate with incomplete information. To mitigate risks in the use of algorithms in decision-making, we consider the question of modeling of beliefs. We argue that evidence supports that beliefs are not always classical statistical variables and they should, in the general case, be considered as superposition states of disjoint or polar outcomes. We analyze the disjunction effect from the perspective of the belief as a quantum vector.

**Introduction**
It is well known that judgment based on probability computations can be erroneous if the experiment is not properly designed or if the data is inconsistent due to incompatible sample sizes of the components or the presence of a confounding variable. An example of this is the Simpson's paradox [1], where the association between two variables (X, Y) reverses sign upon conditioning of a third variable Z, regardless of the value taken by Z (e.g. [2]). If comparisons between two different groups point to one judgment and which reverses when the groups are aggregated, how is one to reason in a sensible manner?

Given that E and C are effect and cause variables, the general idea in probabilistic reasoning is that P(*E*|*C*) > P(*E*|*C'*) where *C'* represents the absence of the cause event. But Pearl has argued [2],[3] that to establish causality one must show that the cause leads to the effect in an active setting and not as a consequence of a passive correlation. He asserts that a causal relationship can be established only when P(*E*|*do(C)*) > P(*E*|*do(C')*), where *do(C)* stands for an external intervention that compels the truth of the causation.

The data used to establish correlation in a complex system does not always admit testing for causation [4]. This has significant consequences in reasoning systems and sometimes incorrect or dubious causation may be accepted. In applications in an artificial intelligence scenario, programs working with disaggregated or aggregated data may arrive at different judgments, and so the use of algorithms in decision-making comes with risks that must be mitigated by a proper understanding of the reasoning process. Since decisions on how to aggregate data are made by human agents and written into the mathematics that constitutes the algorithm, it is also essential to understand the structuring of belief in the human mind.

When considering counter-intuitive effects, the question arises as to how beliefs are handled by the mind. It is customary to take the mind as a classical system, where belief is a classical variable, in accord with the mainstream understanding of the functioning of the brain. In use of this approach to reasoning, the statistics obtained about a property true of a population are associated with each member of the set and members of similar groups that we might interact with in the future.

Reasoning may also be seen through the lens of quantum decision theory, which is being increasingly applied to real world problems. Based on the mathematics of Hilbert spaces, this formalism has the capacity to capture aspects of cognitive and social processing that are not accessible to classical models. In recent years, experiments and theory both have come out on the side of quantum biology [5], with quantum mechanics playing a direct or indirect role in photosynthesis [6], olfaction, vision [7], bird navigation and long-range electron transfer. Although, there is no consensus on models of quantum processing in the brain, there is substantial support for the view that such models are appropriate for cognition [8].

Quantum languages appear to belong to the language hierarchy associated with the brain [9],[10], so it is possible that "human-like" processing of sensory and actuator information involves consideration of latent variables has a quantum mechanical basis. Quantum models (e.g. [8]) have been used to explain counter-intuitive data concerning conjunction and disjunction fallacies from behavioral psychology where real choices sometimes violate the expected utility hypothesis.

In this paper, we interpret some experimental data to suggest that beliefs are characterized by "wholeness." We begin with the problem of probability reversal and examine its relationship to the workings of the reasoning process. Specifically, we argue against the representation of belief as a classical variable and present evidence in favor of it being a superposition state that is best handled using quantum logic. We use this approach to provide a new take on the disjunction effect.

**Simpson's Paradox**
An example of Simpson's paradox is provided by real medical data [1] of Table 1, where different sets of individuals suffering from a specific illness are subjected to two different treatments. Treatment 1 is superior to Treatment 2 for each of the two studies. But if the two studies are aggregated then Treatment 2 is superior. If the two treatments are termed A and B and study represented by a subscript, then as far as beliefs are concerned $P(A_i) > P(B_i)$ for $i =1,2$, but $P(B_1) + P(B_2) > P(A_1) + P(A_2)$.

Table 1. Simpson's probability reversal

|  | Study 1 | Study 2 | Totals |
|---|---|---|---|
| Treatment 1 | 0.93 (81/87) | 0.73 (192/263) | 0.78 (273/350) |
| Treatment 2 | 0.83 (234/270) | 0.69 (55/80) | 0.83 (289/350) |

As far as the two studies go, there is much variation in the design. The sample sizes of the two treatments vary with a proportion of 3:1 and this is a contributing factor to the probability reversal.

This paradox arises from the fact that if $a_1/b_1 > c_1/d_1$ and $a_2/b_2 > c_2/d_2$ there will be situations so that $(c_1+c_2)/(d_1+d_2) > (a_1+a_2)/(b_1+b_2)$. Its working in our mind arises from the fact that we consider each judgment separately as an integral whole without caring for the specific numbers associated with it.

Pearl argues that accounting for the role of causality resolves the paradox [2]. He speaks up for the use of a causal calculus to determine which set of contradictory findings should be accepted, and he argues for disaggregating the data by conditioning it on a third variable that is causally related. However it may not be possible to perform such causal calculus in an application that is running in real time.

A more dramatic example, which comes with two conditioning variables, is given in Table 2 (see [11]). This consists of different sets of patients separated by gender (separate studies for males and females), where some patients are given the treatment and others are not (who constitute the control group).

Table 2. Treatment figures separated by gender

|  |  | Improvement | No Improvement | Totals | % |
|---|---|---|---|---|---|
| Males | Treatment | 18 | 12 | 30 | 60% |
|  | Control | 7 | 3 | 10 | 70% |
| Females | Treatment | 2 | 8 | 10 | 20% |
|  | Control | 9 | 21 | 30 | 30% |
| Totals | Treatment | 20 | 20 | 40 | 50% |
|  | Control | 16 | 24 | 40 | 40% |

As we can see, the control group shows better results both for males and females with the spread in percentages of 70 to 60 and 30 to 20 (the total samples are in a proportion of 3:1 in the two studies). In other words, "no treatment" is a better option than "treatment" for both males and females. When the studies are combined, a reversal takes place and "treatment" turns out to be better than "no treatment" with a proportion of 50 to 40 percent. One could take this to imply that if the sex of the patient is known then "no treatment" is better, but if it is unknown then "treatment" is better!

The human mind is quite satisfied with the conclusions made from the disaggregated data since it forms an integral part of the narrative at the level of the gender-wise treatment.

**Two-stage Experiments**

Now consider a treatment protocol that is performed in two stages. In one design meant to determine the effect of choice, all patients are randomly assigned to one of two groups: the random group and the option group to see the effects of self-selection bias [12]. Other designs test for the mutual relationship between different treatments.

In the design we discuss now, the patients in the test are given two treatments in stages and these are A and B in any of the four combinations $A_1A_2$, $A_1B_2$, $B_1A_2$ $B_1B_2$ (B might be a placebo but that detail is not important to our discussion). The total number of patients is divided into four groups and their assignment and numbers are determined by clinical factors. The improvement numbers are given in Table 3. These may be seen as the values at nodes at the end of a two-stage tree.

Table 3. A two-stage treatment regime, with fraction of patients who showed improvement

|  | Second stage $A_2$ | Second stage $B_2$ |
|---|---|---|
| First stage $A_1$ | 0.80 (36/45) | 0.20 (3/15) |
| First stage $B_1$ | 0.10 (1/10) | 0.30(9/30) |

The improvement numbers are normalized in Table 4, where we have divided the actual improvement fractions by the total, and these may be considered the probability estimates for the nodes at the bottom of the two-stage tree.

Table 4. Normalized percentages of all improved patients

|  | Second stage A | Second stage B |
|---|---|---|
| First stage A | 0.57 | 0.14 |
| First stage B | 0.07 | 0.22 |

We do not look into the question of if statistical tests (such as Fisher's Exact Test) to show the degree of dependence between the four cases. If we add up the two rows of Table 4, we get P(A) = 0.71 and P(B) = 0.29. The treatments in the two stages are not independent since P(AA) ≠ P(A)$^2$, P(BB) ≠ P(B)$^2$ and so on. The order of the treatment matters. Since P($A_1B_2$)≠P($B_1A_2$), the two treatments cannot be considered to be events A and B of Figure 1 in a two stage repeated experiment. The order effect may be rewritten as P(AB) ≠ P(BA).

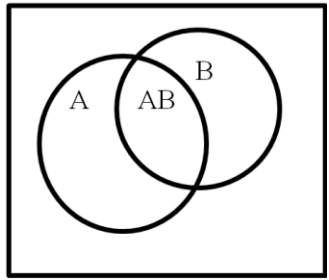

Figure 1. Two-stage treatment regime

The example of Table 3 for two-stage treatments is similar to responses to two successive questions that test the individuals' beliefs.

**Belief vectors**

Now consider the question of how beliefs, which are relations between states, are maintained? In general one can speak of a belief vector that underlies the reasoning process. A specific belief may be a numerical value or a Boolean variable that represents a choice between polar opposites (a dichotomy).

A belief as a random variable B may also be seen in relation to a small set of variables that directly affect its value in the sense that B is conditionally independent of other variables. The set of locally affecting random variables forms a part of the belief network which is a directed model of conditional dependence among the variables. Effectively, the belief network provides a factorization of the joint probability distribution connecting the belief variable to other variables. Reasoning may be based on latent variables or others that are not easily accessed. In certain cases, conditional dependencies could be determined by other secondary variables but this may be prohibitively expensive.

Although we program machines to arrive at clear judgments, this is different from how humans think. Psychologists tell us that humans do not have an absolutely black-and-white cognitive style. Indeed, dichotomous thinking is a characteristic of borderline personality disorder [13]. One may conclude that dichotomous beliefs represent a "fluid" position that changes in jumps based on various internal and external states. Quantum logic sidesteps the question of frozen choice by speaking of superposition of opposites. The superposition vector rotates in an abstract space and this rotation accords with the intuition that our reasoning has a geometric side to it. The actual choice made is probabilistic and it equals the projection of the superposition vector on the cognition axis.

In classical physics, the state of a system of mass m is shown by a point (p, q) in the phase space of positions q and momenta p. Physical magnitudes are represented by real functions that commute and understood to have definite values at any time, irrespective of whether they are observed or not. Physical events can be related using set theoretical operations such as intersection and union and this gives rise to a Boolean algebra.

In quantum theory, the physical system is represented as a complex separable Hilbert space H, with a pure state represented by a ray in H. Physical magnitudes are mappings of self-adjoint operators on H that do not commute under multiplication. This noncommutavity means that quantum magnitudes cannot be considered to be pre-existent prior to observation [14].

In classical logic, the distributive law of propositional logic is valid for the variables are clearly defined. Therefore, observation O may be written as the proposition $O \cap (Path1 \cup Path2)$, which by distribution equals $(O \cap Path1) \cup (O \cap Path2)$. Not so in quantum logic, where we can only speak of the observation and not a reality that underlies that observation. For example, analysis of an observation may reveal that a particle can be taken to have traveled simultaneously through two paths (as in the Mach-Zehnder interferometer).

Similar to the particle's passage through two different paths in quantum theory, beliefs can simultaneously encompass mutually contradictory aspects. For example, a popular position in certain academic circles in the United States after the economic downturn of 2008 was that a big stimulus was needed for economic growth, yet the same people averred that caring for the environment required that growth be slowed. There were others who said that for lessening worldwide inequality, the poor countries should be given incentives to grow their economies, but they also said this can only happen if the rich countries increased consumption (which exacerbates the inequality). Some others, whose focus is social change, were concerned that there were too many people making less than the average, so there was need for the economy to lift people above the average to ensure this was no longer true!

At a personal level, the pulls of Eros and Thanatos are the basis of human personality and these oppositions are the ground on which many beliefs are formed. One would, therefore, find contradictions underlying many beliefs, especially when they are examined thoroughly.

A Gallup Poll in 1997 asked 1,002 respondents if Bill Clinton was honest and trustworthy with a response of 50%, with the follow up question asking the same about Al Gore came up with a response of 60%. When the order of the questions was reversed, the responses were 68% and 57%, respectively [15]. Various kinds of order effects in surveys have been noted and they have been classified into different types and various quantum cognition explanations have been furnished for them [8],[16].

The proposal that mental processing is governed by quantum mechanics was made long ago by Bohr [17] and it is important to note that this is not contingent on the brain working as a quantum machine [18],[19]. The idea here is that the abstract space associated with memories and beliefs supports coherent processing based on quantum logic. This abstract space is not taken to be equivalent to the activity in the brain and it is seen as emergent on the brain processes, just as chemistry is emergent on physics and biology is emergent on chemistry. The order effects are naturally explained by considering the variables to be quantum states.

According to the orthodox interpretation of quantum theory, cognitive functions cannot be explained using any ontology based on physical structures [20]. The consideration of quantum cognition opens up the possibility that the quantum resource of entanglement underlies paradoxical behavior of human agents in social networks and, in turn, it could be exploited to provide superior performance in engineered and business systems.

The quantum belief state associated with the two-stage experiment of Table 3 is (where the probability amplitudes are the square-root of the corresponding probabilities of Table 4):

$$|\varphi\rangle = 0.75|AA\rangle + 0.37|AB\rangle + 0.26|BA\rangle + 0.47|BB\rangle$$

As a quantum tree [21], this may be mapped to Figure 2, where the changing branching probabilities across the stages are a consequence of the entanglement between the variables.

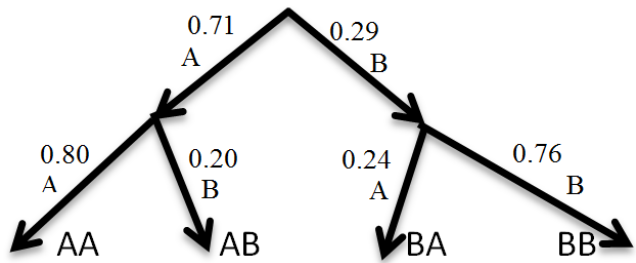

Figure 2. Quantum tree probabilities for the two-stage example

The tree representation of Figure 2 provides a clear idea of the nature of the order relation between the two treatments. This structure can be further enlarged if additional alternation of treatment is a part of the design. The final state is a mixed state that cannot be, in general, viewed as a product of elementary states.

**The Disjunction Effect**

Mathematically, the disjunction effect is the false judgement that the probability P(A or B) is less than either P(A) or P(B). It changes cognitive judgment based on information with no computational value.

Consider the setting of a gamble (with purported 50% chance of win) in which a win is worth $200 and a loss makes one lose $100. If told that they had won the first gamble, 69% decided to play again; told that they had lost, 59% chose to play again. But only 36% chose to accept the second gamble if they were not told the outcome of the first [22],[23]. In other words, the decision supports the incorrect P(A) > P(A or B).

There are many explanations of this effect, including prospect theory [24] in which decisions are ordered in an initial editing phase according to a certain heuristic. In particular, people decide which outcomes they consider equivalent, establish a reference point and then consider lesser outcomes as losses and greater ones as gains. In the subsequent evaluation phase, people compute a utility, based on the potential outcomes and then choose the alternative having a higher utility.

We believe that the most satisfactory explanation of the disjunction effect is to view the beliefs about the outcome as a quantum version of prospect theory (Figure 3). Since the Quantum Zeno Effect (e.g. [25]) freezes the state upon observation, the odds do not change after the results of the first play are known and, therefore, the players play a second time irrespective of whether they have won or lost. If the outcome of the first gamble is not revealed, the belief state evolves (rotates further) where the odds of losing become larger than that of winning.

It seems reasonable to take the reference point as the superposition state $\sqrt{\frac{2}{3}}|1\rangle + \sqrt{\frac{1}{3}}|2\rangle$ where $|1\rangle$ is the outcome of losing $100 is and $|2\rangle$ is the state of winning $200. This balances the two outcomes for the probabilities of the two outcomes are 2/3 and 1/3, respectively, with a utility of $-2/3 \times 100 + 1/3 \times 200 = 0$.

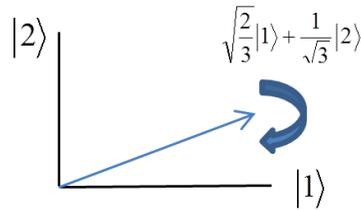

Figure 3. The prospect as a quantum state

The evolution of the state is guided by the same intuition that is at the basis of the St. Petersburg Paradox [26]. In this paradox, originally proposed in 1738, the house offers to flip a coin until it comes up heads. The house pays $1 if heads appears on the first trial; otherwise the payoff doubles each time tails appears, with the compounding stopped and payment given at the first heads. Although this gamble has an infinite expected value, most people agree that they should not offer more than a few dollars to play. Daniel Bernoulli offered an explanation based on the marginal value of money to the individual diminishing as the wealth rises. Another factor in the explanation is that players have a limited amount of money to play and he could be wiped out before he has made the killing.

The arrow of the reference state in Figure 3 will turn clockwise as the gamble is played successively, without knowing the outcome. On the other hand, the knowledge of the outcome resets the reference at its original location where the utility is zero. The use of quantum logic in the resolution to the counter-intuitive behavior of rational agents supports the view that the abstract space in which beliefs are held supports coherence.

**Conclusions**

Data based judgments as part of reasoning play a fundamental role in cognition. In some situations, such as Simpson's reversal and the disjunction effect, these judgments come with paradoxical reversal when data is aggregated at different levels. In order to understand risks in the use of algorithms in decision-making, we examined how beliefs might be processed as statistical and quantum variables.

The use of quantum logic appears to provide new insight into the workings of agents in different decision environments. The belief vector is a superposition vector of mutually exclusive outcomes which rotates when it interacts with the vector of internal and external states. This may not provide us a method of predicting behavior, but it can be a useful method of its understanding.